\documentclass[10pt,twocolumn,letterpaper]{article}

\usepackage{iccv}
\usepackage{times}
\usepackage{epsfig}
\usepackage{graphicx}
\usepackage{amsmath}
\usepackage{amssymb}
\usepackage{subfigure}

\usepackage[colorlinks, bookmarks=ture]{hyperref}

\iccvfinalcopy 

\def\iccvPaperID{11883} 

\ificcvfinal\fi

\begin{document}

\title{DEYO: DETR with YOLO for Step-by-Step Object Detection}

\author{Haodong Ouyang\\
Southwest Minzu University\\
Chengdu, China\\
{\tt\small ouyanghaodong@stu.swun.edu.cn}
}

\maketitle
\ificcvfinal\thispagestyle{plain}
\setlength{\textfloatsep}{0.5pt}

\begin{abstract}
   Object detection is an important topic in computer vision, with post-processing, an essential part of the typical object detection pipeline, posing a significant bottleneck affecting the performance of traditional object detection models. The detection transformer (DETR), as the first end-to-end object detection model, discards the requirement of manual components like the anchor and non-maximum suppression (NMS), significantly simplifying the object detection process. However , compared with most traditional object detection models, DETR converges very slowly, and a query’s meaning is obscure. Thus, inspired by the Step-by-Step concept, this paper proposes a new two-stage object detection model, named DETR with YOLO (DEYO), which relies on a progressive inference to solve the above problems. DEYO is a two-stage architecture comprising a classic object detection model and a DETR-like model as the first and second stages, respectively. Specifically, the first stage provides high-quality query and anchor feeding into the second stage, improving the performance and efficiency of the second stage compared to the original DETR model. Meanwhile, the second stage compensates for the performance degradation caused by the first stage detector’s limitations. Extensive experiments demonstrate that DEYO attains 50.6 AP and 52.1 AP in 12 and 36 epochs, respectively, while utilizing ResNet-50 as the backbone and multi-scale features on the COCO dataset. Compared with DINO, an optimal DETR-like model, the developed DEYO model affords a significant performance improvement of 1.6 AP and 1.2 AP in two epoch settings.
\end{abstract}

\section{Introduction}
Object Detection involves a process where a detector identifies the region of interest within an image and marks it with a bounding box and a class label. After years of vigorous advancement of the classic object detectors, several excellent one-stage and two-stage object detection models have been developed. Modern detectors usually comprise
\begin{figure}[h]
\begin{center}
\includegraphics[width=0.75\linewidth]{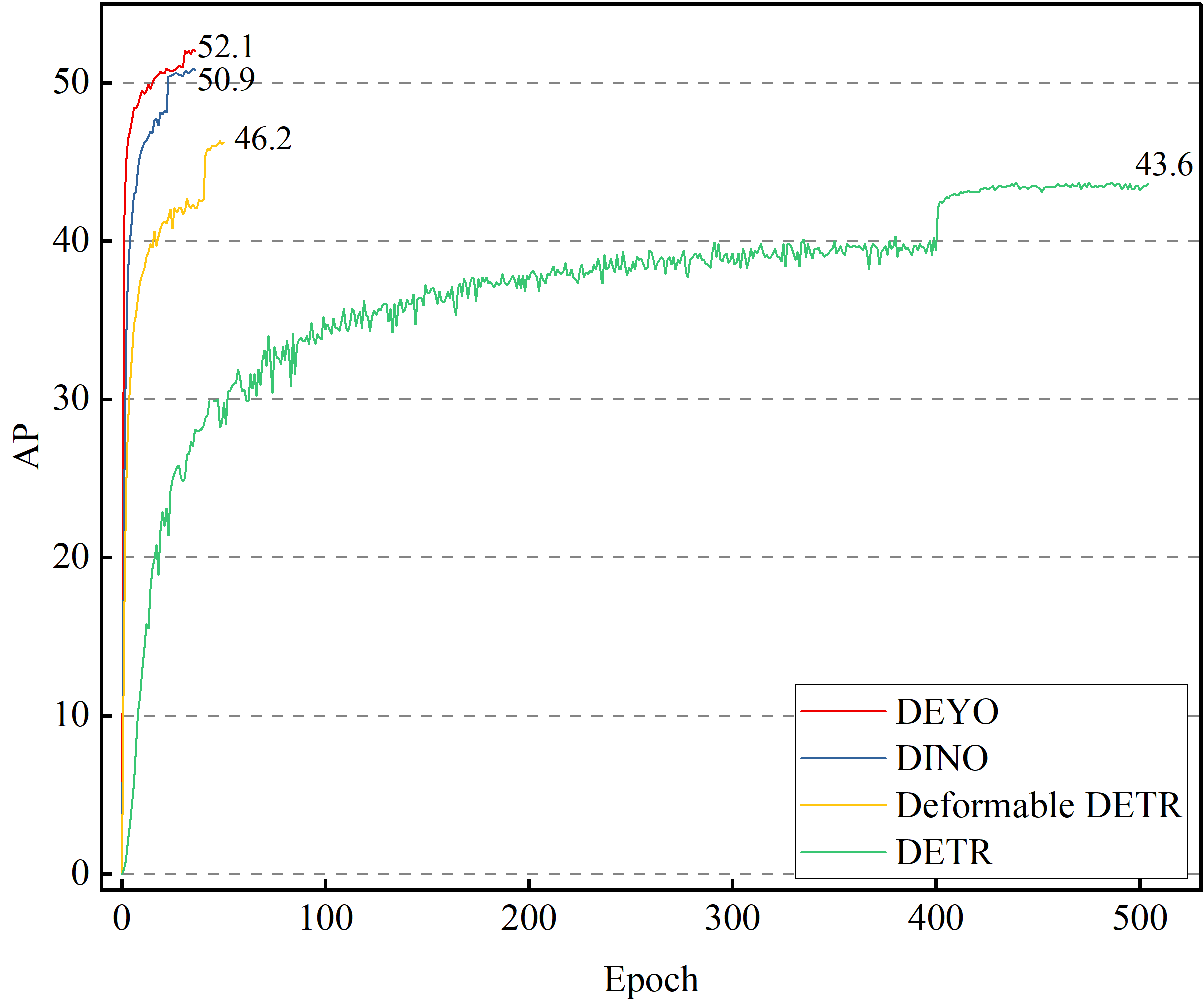}
   \caption{Convergence curves between DEYO and previous models under ResNet-50 backbone.
}
\end{center}

\end{figure}

\noindent two parts, i.e., a backbone and a head for
predicting the object classes and bounding boxes. Recent architectures insert layers between the backbone and the head to collect feature maps from different stages. These layers are referred to as the object detector’s neck. The R-CNN series \cite{36} is the most representative two-stage object detector, including Fast R-CNN \cite{16} and Faster R-CNN \cite{9}. The most representative one-stage object detector models are YOLO \cite{4,5,6}, SSD \cite{20}, and RetinaNet \cite{21}. A common feature of these classic object detectors is heavily relying on handcrafted components, such as anchor and non-maximum suppression (NMS), with NMS being a post-processing filter removing redundant bounding boxes. In recent years, anchor-free methods, such as CenterNet \cite{17}, CornerNet \cite{18}, and FCOS \cite{19}, have achieved comparable results with the anchor-based models. However, both schemes, anchor-based and anchor-free, utilize non-maximum suppression for post-processing, posing a bottleneck to classic detectors. Additionally, since non-maximum suppression does not use image information, it is error-prone in bounding box retention and deletion.

Classic detectors are mainly based on convolutional neural networks. Carion et al. \cite{1} proposed an end-to-end detector called DETR (DEtection Transformer), which relies on the transformer-based encoder-decoder architecture \cite{10} to eliminate the dependence on anchor and NMS manual components. Moreover, DETR uses the hungarian loss to directly predict the one-to-one object set, simplifying the object detection pipeline. Although DETR has many attractive features, it suffers from several problems. First, DETR requires 500 training epochs to achieve an appealing performance. Moreover, the DETR’s query is obscure, prohibiting its full utilization. It should be noted that a series of DETR-based variant models have been proposed, which tackled the above problems well. For instance, by designing a new attention module, the deformable DETR focuses on the sampling points around the reference point to enhance the cross attention’s efficiency. The Conditional DETR \cite{11} decouples the DETR query into content and a location part to clarify the query’s meaning. Additionally, DAB-DETR \cite{8} considers the query a 4D anchor box and improves it layer by layer. Based on DAB-DETR, DN-DETR \cite{13} suggests that the slow convergence of DETR’s training is due to the bipartite graph matching instability in the early training stage. Hence introducing denoising group technology significantly accelerates the DETR model convergence. Spurred by the above models, DINO \cite{14} has further improved DETR by employing Object365 \cite{37} for detection pre-training and a Swin-Transformer \cite{23} as the backbone network. Indeed, DINO attained a state-of-the- art (SOTA) result of 63.3AP on the COCO val2017 dataset \cite{12}. Currently, DINO has the fastest training convergence speed and the highest accuracy among the DETR-like models and proves that DETR-like detectors achieve an equal performance or even outperform the classic detectors.

Although relevant works have significantly improved the DETR-like models, we still consider that it is difficult for DETR-like models to directly predict the one-to-one object sets. Classic detectors, such as YOLOv5 \cite{24}, can generate 25,200 predictions as its output in a 640x640 pixels image. Given that the Transformer model calculations used by DETR increase squarely with the number of queries, the number of queries of the DETR-like model is typically in the 100 to 900 range. Obviously, a single YOLO prediction has a much lower computational burden than a DETR-like model. Hence, inspired by the Step-by-Step idea, we use the low-cost and high-quality YOLO predictions as inputs to the second-stage DETR-like model to reduce the difficulty level of predicting the one-to-one object sets. This strategy affords the limited number of queries of the DETR-like model to focus on challenging tasks, such as objects that are difficult to recognize and heavily occluded objects, improving the overall performance.

This paper proposes a new two-stage object detection model using progressive inference. Specifically, our model utilizes YOLOv5, an advanced real-time object detection model, as the first stage and a DETR-like model as the
second stage. The output of the YOLO model is processed by the transition components, comprising the object and bounding box information, which are then passed to the DETR-like model. The high-quality initialization query of YOLO, the initialization queries of anchor and DETR, and the initialized anchor are combined and then sent to the Transformer's decoder. The experimental results demonstrate that the second stage decoder of the developed model can easily recognize the information from the first stage. Indeed, the DETR-like models focus more on fine- tuning the initial bounding box, verifying and adjusting the categories, and predicting the objects that NMS erroneously filters out due to severe occlusions or objects that the first stage detector cannot easily recognize. Additionally, due to the existence of the high-quality initialization queries and anchors, the optimization goal is clarified for the initialization of both queries and anchors in the second stage. The instability of bipartite graph matching in the second stage is further reduced, and training time is greatly accelerated.
The proposed two-stage network is mutually reinforcing, as the model’s first stage provides high-quality initialization for the second stage so that the latter stage can quickly focus on difficult-to-learn information. This concept accelerates the DETR-like models’ convergence and improves their peak performance. The second stage model fine-tunes the first stage model to achieve better outcomes. Hence, the proposed model compensates for the performance degradation of the classic detectors due to the NMS limitations affording the model to identify severely occluded objects.

To our knowledge, this is the first work introducing progressive inference into a DETR-like detection model with using classic object detectors . Our contributions are summarized as follows:

1. Designing a new two-stage model inspired by the Step-by-Step idea. The experimental results demonstrate that this progressive inference significantly reduces the difficulty level in predicting one-to-one object sets. Moreover, the training convergence time of the DETR-like models is significantly reduced from a novel perspective, while the model’s accuracy is enhanced to a new level.

2. Overcoming the bottleneck performance problem that classic detectors suffer due to NMS. Additionally, we analyze the potential performance of classic detectors after resolving the NMS-based performance bottleneck problem.

3. Conducting several experiments verifying our ideas and exploring the contribution of each component within our model.
\hspace*{\fill}

\section{Related Work}
\noindent{\bf YOLO.} Over the years, the YOLO \cite{4,5,6} series has been one of the best single-stage real-time object detector classes. YOLO detectors can be found in many hardware platforms and application scenarios, meeting different requirements. After years of development, YOLO has evolved into a series of fast models achieving good performance. The anchor-based YOLO methods include YOLOv4 \cite{25}, YOLOv5 \cite{24}, and YOLOv7 \cite{26}, while the anchor-free methods are YOLOX \cite{27}, YOLOR \cite{28}, and YOLOv6 \cite{29}. Considering the performance of these detectors, the anchor-free methods perform equally well as the anchor-based methods, with the anchor no longer being the main factor restricting the development of YOLO. However, all YOLO variants suffer from generating many redundant bounding boxes that NMS must filter out in the prediction stage, which can lead to performance bottlenecks. In our model, this problem is ameliorated to a certain extent.

\hspace*{\fill}

\noindent{\bf Non-Maximum Suppression (NMS).} NMS is a significant component of the classic object detection pipeline, aiming to select the best bounding box from a set of overlapping boxes. NMS sort all bounding boxes according to their scores. Select the highest scoring bounding box M and suppress all other bounding boxes that have overlap with M above a threshold value. In recent years, several works attempted to improve NMS, such as Soft-NMS \cite{30}, Softer NMS \cite{31}, and Adaptive NMS \cite{32}. However, none of these can overcome the intrinsic problem of NMS, i.e., filtering out redundant bounding boxes does not consider image information. A major issue with non-maximum suppression is that it sets the score for neighboring detections to zero, if an object was actually present in that overlap threshold, it would be missed and this would lead to a drop in average precision. Therefore, the model’s generalization ability is refrained to a certain extent, affecting the model’s performance on complex tasks.

\hspace*{\fill}

\noindent{\bf DETR and its variants.} Carion et al. \cite{1} proposed an end- to-end object detector based on Transformer named DETR (DEtection TRansformer). DETR has attracted the researchers’ attention due to its end-to-end object detection characteristics. Specifically, DETR removes the anchor and NMS components in the traditional detection pipeline, adopts the label assignment method of bipartite graph matching, and directly predicts the one-to-one object set. This strategy simplifies the object detection process and alleviates the performance bottleneck problems caused by NMS. Additionally, introducing a Transformer architecture makes the bounding boxes filtering interactive between the image and object features, i.e., the object filtering of DETR combines image information, affording DETR to retain and delete boxes correctly. However, DETR suffers from slow convergence and obscure queries. Many DETR variants, such as Conditional DETR \cite{11}, Deformable DETR \cite{7}, DAB-DETR \cite{8}, DN-DETR \cite{13}, and DINO \cite{14}, have been presented to solve these issues. For instance, the Conditional DETR decouples the query into the content and location parts, while the Deformable DETR improves the cross-attention efficiency. DAB- DETR interprets the query as the 4-D anchor boxes and learns to improve them layer by layer. DN-DETR, based on DAB-DETR, introduces a denoising group to solve unstable bipartite graph matching and significantly accelerates the model training convergence. DINO is a DETR-like model that further improves the previous works and achieves SOTA results. \cite{43} proposed to introduce RetinaNet\cite{21}-like model into a sparse R-CNN \cite{46} to improve the performance of the detector.

\begin{figure}[t]
\begin{center}
\includegraphics[width=0.8\linewidth]{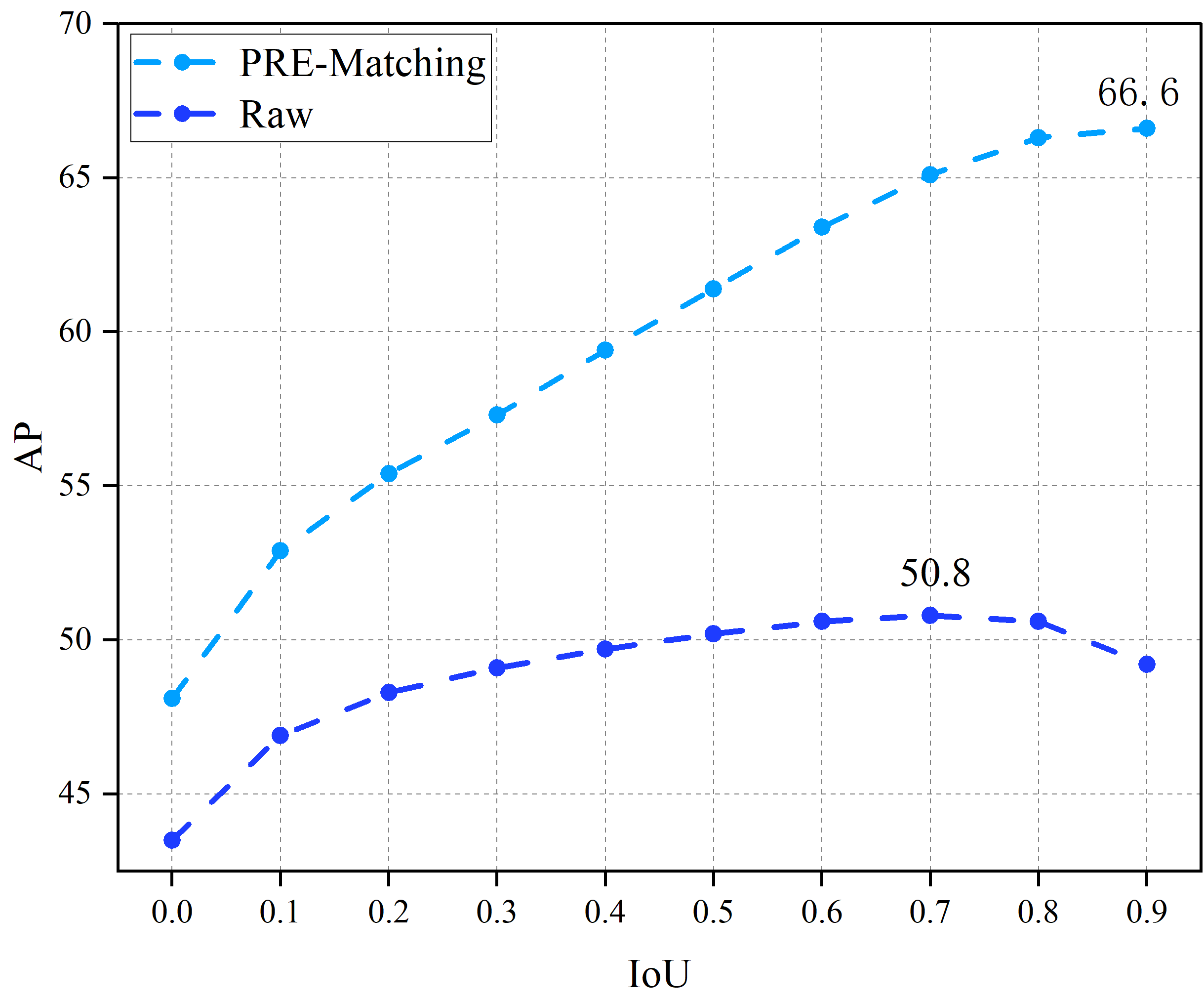}
   \caption{YOLOv5x compares the performance of the raw output using NMS with that using PRE-Matching, where the purpose of PRE-matching is to find the prediction that best matches the ground truth in the set of predictions filtered by NMS, thus detecting whether the NMS correctly removes the bounding box.
}
\end{center}
\label{fig:2}
\hspace*{\fill}
\end{figure}

\hspace*{\fill}

\noindent{\bf Let’s think step by step.} By using the specific prompt "Let's think step by step" and the corresponding two-stage prompt technique, \cite{42} has improved the inference power of the large-scale language model on several inference- related zero-shot tasks, outperforming the previous zero- shot approach. The let's think step-by-step scheme encourages the model to reason step-by-step for complex tasks where it is difficult to provide the correct answer directly and enables the model to compute an answer that would otherwise not be given correctly. Inspired by the Step-by-Step concept, our model’s first stage predicts simple tasks instead of the final inference, similar to the "intermediate inference step" in a language model. The second stage focuses on the difficult tasks given in the first stage from the "intermediate inference step" and reduces the difficulty, allows the model to predict objects that would otherwise objects that could not be correctly identified.

\hspace*{\fill}
\section{Why is Step-by-Step effective?}
\noindent{\bf High-quality query initialization speeds up training.}
DETR employs the Hungarian matching algorithm to predict the one-to-one object sets directly. However, there is evidence that Hungarian matching does not lead to stable matching due to blocking pairs \cite{3}. The final matching result will change significantly due to the small changes in the cost matrix, leading to inconsistent optimization goals of the decoder query during the early stage of training. This increases the training difficulty remarkably, which is one of the critical reasons for the slow convergence of DETR. Therefore, we input high-quality queries and Anchors into the decoder. As illustrated in Figure~\ref{fig:4}(b), our model achieves 40.6 AP in the first epoch, demonstrating that the decoder can easily learn to obtain information in high-quality queries and anchors, assisting the model in clarifying its optimization goals and avoiding the ambiguity caused by the Hungarian matching. Thus, the slow training convergence due to unstable Hungarian matching is
significantly reduced.

\begin{figure*}[t]
\begin{center}
\includegraphics[width=0.75\textwidth]{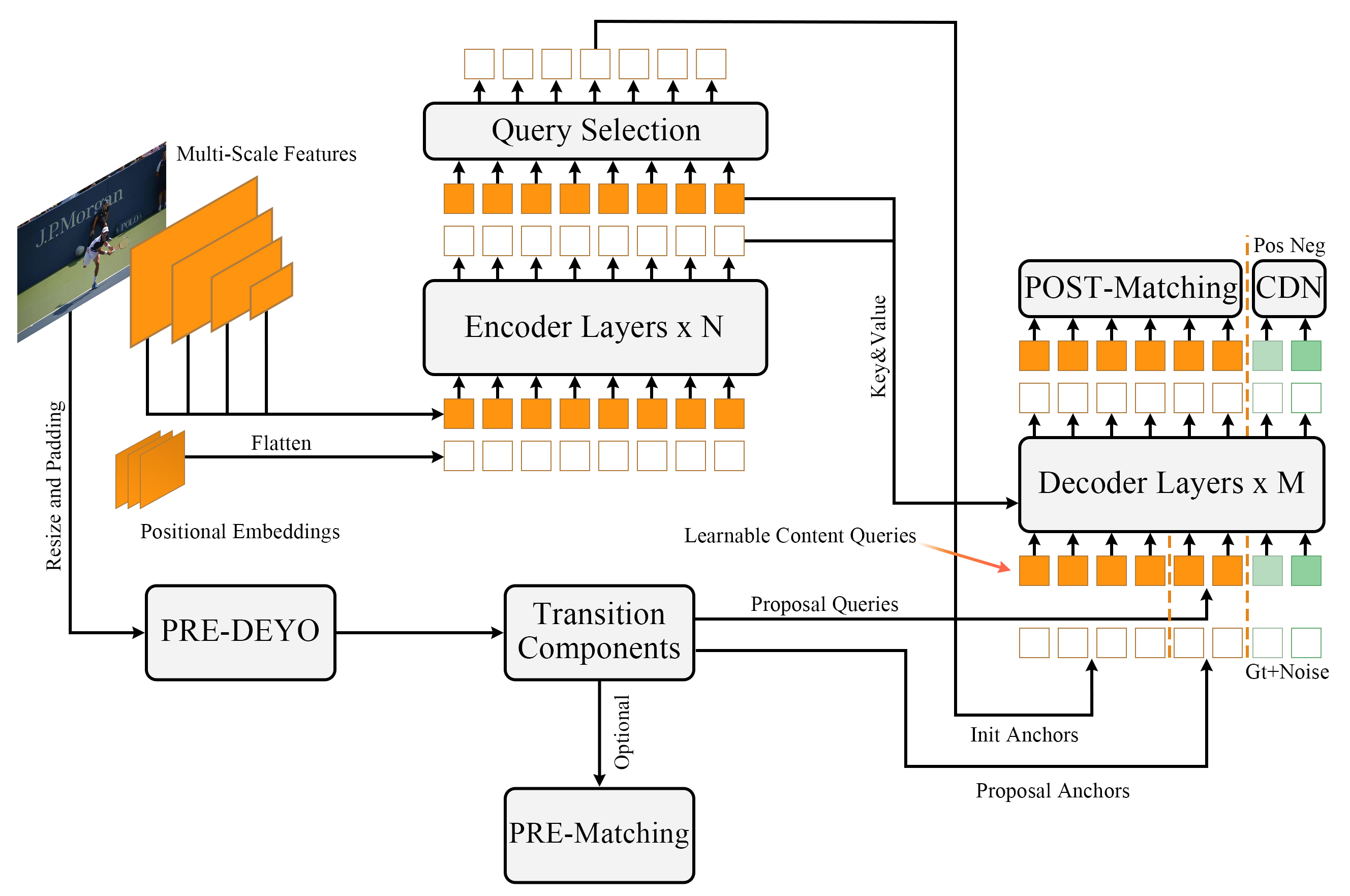}
   \caption{The framework of our DEYO model. DEYO is a progressive inference object detection model, the high-quality proposal of PRE- DEYO feeds into the Transformer decoder of POST-DEYO, and object is predicted step-by-step.
}
\end{center}
\label{fig:3}
\end{figure*}

\noindent{\bf Breaking through the performance bottleneck caused by NMS.} In the prediction stage, the classic detectors generate redundant bounding boxes that must be suppressed and filtered out by NMS. However, the NMS filtering algorithm does not integrate image information but tries to optimize filtering by adjusting the different IoU thresholds according to different tasks, which is error- prone in the retention and deletion of boxes. Hence, we investigated the potential performance of YOLOv5x \cite{24} without using NMS . In Figure~\ref{fig:2}, shows the post-processing performance of YOLOv5 at 640x640 size images with using NMS with different IoU thresholds. The results highlight that the prediction performance using pre-matching filtering after NMS will gradually improve as the IoU threshold increases. Additionally, the prediction performance without pre- matching filtering will gradually decrease as the IoU threshold increases. This observation indicates that in a difficult task, a low threshold can lead to the wrong removal of boxes, while a high threshold can generate redundant boxes and affect the final results. This implies that even if the detection model’s prediction is good, the final results can still be affected by NMS, thus imposing a performance bottleneck. Figure~\ref{fig:2} also depicts the potential performance of YOLOv5x without NMS.

\noindent{\bf One-to-many with one-to-one.}One problem with one-to-one tag assignment is that performance degrades as the number of queries increases when a certain number of queries is reached, i.e., peak performance of one-to-one tag assignment cannot be achieved at the maximum number of queries, i.e., peak performance with one-to-one label assignment is not achieved at the maximum number of queries \cite{44, 45}. DEYO solves this problem by combining the advantages of one-to-many label assignment and one-to-many label assignment, with PRE-DEYO using one-to-many matching. PRE-DEYO uses one-to-many matching to select the high-quality 100 queries from more than 20,000 queries to feed into POST-DEYO, which makes the detection pressure of POST-DEYO drop significantly and allows a limited number of queries to focus on difficult objects that account for a small portion of the total objects.

\section{DEYO}
\subsection{Overview}
Our model uses YOLOv5 as the first stage and DINO \cite{14} as the second stage, affording a new, progressive inference-based two-stage model. In this paper, the first stage YOLOv5 model is referred to as the PRE-DEYO, and the second stage DINO model is the POST-DEYO. As a detector of the classic YOLO series, PRE-DEYO contains a backbone, a neck comprising FPN \cite{34} + PAN \cite{35}, and a head that outputs three-scale prediction information. As a DETR-like model, POST-DEYO contains a backbone, a multi-layer Transformer encoder \cite{10}, a multi-layer Transformer decoder \cite{10}, and multiple prediction heads. It uses static query and dynamic initialization of the anchor boxes and involves an additional CDN branch for comparative denoising training. The overall DEYO model is illustrated in Figure~\ref{fig:3}. The output of PRE-DEYO is combined with the initialization query and anchor of PRE-DEYO through the transition components, which are input to the Transformer decoder. The POST-DEYO can quickly acquire the information of PRE-DEYO during training and focus on difficult tasks. We give two label assignment methods, DEYO uses POST-Matching by default, PRE- Matching is only used in ablation study. PRE-Matching is similar to POST-Matching, the output of PRE-DEYO is directly matched with ground truth to avoid the effects of dichotomous matching instability on target assignment.
\subsection{DINO briefing}
DINO is a DETR-like model based on DN-DETR \cite{13}, DAB-DETR \cite{8}, and Deformable-DETR \cite{7}, formulating queries in the decoder as dynamic anchor boxes and progressively refines them in the decoder layer. After DN- DETR, DINO improves DeNoising Training to Contrastive DeNoising Training (CDN), increasing its ability to predict "no-object" for anchors with no nearby objects while stabilizing the bipartite graph matching during training. Meanwhile, DINO also uses Deformable Attention to improve its computing efficiency. The layer-by-layer refinement of dynamic anchor boxes helps POST-DEYO fine-tune high-quality anchor boxes from PRE-DEYO in inference. Deformable Attention combined with high- quality anchor boxes enables POST-DEYO to quickly find key information in images, further accelerating the bounding box filtering and verifying and adjusting image contents.
\subsection{Transition components}
The transition component processes the information sent to POST-DETR from PRE-DEYO to ensure consistency during information interpretation and ensures that the filtered information from PRE-DEYO is the most appropriate for POST-DEYO.

\noindent{\bf Prediction selection.} PRE-DEYO predicts many nearly identical bounding boxes, imposing the POST-DEYO to crash during training if no filtering mechanism is introduced. We find that the filtering ability in POST-DEYO is limited. Thus it is very difficult for POST-DEYO to learn the correct filtering strategy for similar and overlapping boxes. Hence, we include NMS in the transition component to filter information from PRE- DEYO. To guarantee the model’s final performance, we obtain high-quality query and anchor boxes that are most suitable for POST-DEYO by adjusting a suitable IoU threshold. Using NMS in the transition component does not limit our model’s final performance because possible performance degradation due to the wrong retention or deletion of boxes is compensated in POST-DEYO.

\noindent{\bf Padding.} Since the number of objects in each image changes dynamically, the high-quality queries and anchors
produced by PRE-DEYO are padded to a specific number before sending them to POST-DEYO. This strategy ensures the stability of the number of queries in each epoch. The padding queries do not participate in bipartite graph matching or loss calculations and are not used in the final prediction results.

\noindent{\bf Label Mapper.} The COCO \cite{12} category indexes of PRE- DEYO ranges from 0 to 79, and the COCO category indexes of POST-DEYO is from 0 to 90, involving several unused category indexes in POST-DEYO. Label Mapper replaces the category index of PRE-DEYO for the sequence number used by the same category of the CDN component in POST-DEYO. Aligning the category indexes allows the model to learn only one type of coding system and accelerate model training.

\noindent{\bf Class Embedding.} The category information of PRE- DEYO is projected to the hidden feature dimension through Class Embedding and is then sent to the Transformer encoder. Since the COCO category number is consistent, our Class Embedding is consistent with CDN's Label Embedding but independent of Label Embedding, significantly speeding up the process where the model learns the PRE-DEYO category information.

\noindent{\bf Post Processing of Anchor.} Since the inference of PRE- DEYO and POST-DEYO is conducted at different image scales, Anchor Post Processing aligns the high-quality anchor of the PRE-DEYO to the scale of POST-DEYO. After normalization and inverse sigmoid processing, the anchor is transferred to the decoder of POST-DEYO.

\begin{table*}[t]
\begin{center}
\begin{tabular}{lccccccccc}
\hline
Model & Epochs & AP & AP$_5$$_0$ & AP$_7$$_5$ & AP$_S$ & AP$_M$ & AP$_L$ & Params & FPS\\
\hline
Faster-RCNN(5scale) \cite{9} &12 & 37.9 &58.8 &41.1 &22.4 &41.1 &49.1 &40M &21$^*$\\
DETR(DC5) \cite{1} &12 & 15.5 &29.4 &14.5 &4.3 &15.1 &26.7 &41M &20\\
Deformable DETR(4scale) \cite{7} &12 & 41.1 &-- &-- &-- &-- &-- &40M &24\\
DAB-DETR(DC5)$^\dagger$ \cite{8} &12 &38.0 &60.3 & 39.8 &19.2 &40.9 &55.4 & 44M &17 \\
Dynamic DETR(5scale) \cite{2} &12 &42.9 &61.0 &46.3 &24.6 &44.9 &54.4 &58M &-- \\
Dynamic Head(5scale)x \cite{41} &12 &43.0 &60.7 &46.8 &24.7 &46.4 &53.9 &-- &--\\
HTC(5scale) \cite{40} &12 & 42.3  &-- &-- &-- &-- &-- &80M &5$^*$\\
DN-Deformable DETR(4scale)$^\dagger$\cite{13} &12 &43.4 &61.9 &47.2 &24.8 &46.8 &59.4 &48M &23\\
DINO-4scale$^\dagger$  \cite{14} &12 &49.0 &66.6 &53.5 &32.0 &52.3 &63.0 &47M &24\\
DINO-5scale$^\dagger$ \cite{14} &12 &49.4 &66.9 &53.8 &32.3 &52.5 &63.9 &47M &10\\
\hline
DEYO-4scale &12 &\bf{50.4} &68.4 &54.9 &33.4 &54.3 &64.9 &132M &12\\
DEYO-4scale$^\dagger$  &12 &\bf{50.6} &68.7 &55.1 &33.4 &54.7 &65.3 &132M &12\\

\hline
\end{tabular}
\end{center}
\caption{Results for DEYO and other detection models with the ResNet50 backbone on COCO val2017 trained with 12 epochs (the so called 1× setting). For models without multiscale features, we tested their FPS for their best model ResNet-50-DC5. DEYO uses 300 queries and 900 queries. $\dagger$ indicates models that use 900 queries or 300 queries with 3 patterns which has similar effect with 900 queries. * indicates that they are tested using the mmdetection \cite{5} framework.}
\label{table:1}
\hspace*{\fill}
\end{table*}

\begin{table*}[t]
\begin{center}
\begin{tabular}{lcccccccc}
\hline
Model & Epochs & AP & AP$_5$$_0$ & AP$_7$$_5$ & AP$_S$ & AP$_M$ & AP$_L$ \\
\hline
Faster-RCNN \cite{9} &108 & 42.0 &62.4 &44.2 &20.5 &44.8 &61.1\\
DETR(DC5) \cite{1} &500 & 43.3 &63.1 &45.9 &22.5 &47.3 &61.1\\
Deformable DETR \cite{7} &50 & 46.2 &65.2 &50.0 &28.8 &49.2 &61.7\\
SMCA-R \cite{38} & 50 &43.7 &63.6 &47.2 &24.2 &47.0 &60.4\\
TSP-RCNN-R \cite{39} & 96 &45.0 &64.5 &49.6 &29.7 &47.7 &58.0\\
Dynamic DETR(5scale) \cite{2} &50 &47.2 &65.9 &51.1 &28.6 &49.3 &59.1\\
DAB-Deformable DETR \cite{8} &50 &46.9 &66.0 &50.8 &30.1 &50.4 &62.5\\
DN-Deformable DETR \cite{13} &50 &48.6 &67.4 &52.2 &31.0 &52.0 &63.7\\
YOLOv5n \cite{24} &-- &28.0 &-- &-- &-- &-- &--\\
YOLOv5x \cite{24} &-- &50.7 &-- &-- &-- &-- &--\\
\hline
DINO-4scale  \cite{14} &24 &50.4 &68.4 &54.9 &34.0 &53.6 &64.6\\
DINO-5scale \cite{14} &24 &51.3 &69.1 &56.0 &34.5 &54.2 &65.8\\
DEYO-4scale &24 &\bf{51.7} &70.0 &56.3 &34.6 &55.6 &65.8\\
\hline
DINO-4scale  \cite{14} &36 &50.9 &69.0 &55.3 &34.6 &54.1 &64.6\\
DINO-5scale \cite{14} &36 &51.2 &69.0 &55.8 &35.0 &54.3 &65.3\\
DEYO-4scale &36 &\bf{52.1} &70.3 &56.6 &34.9 &55.6 &66.0\\
\hline
\end{tabular}
\end{center}
\caption{Results for DEYO and other detection models with the ResNet-50 backbone (except YOLOv5) on COCO val2017 trained with more epochs (24, 36, or more).}
\label{table:2}
\end{table*}


\section{Experiments}
\subsection{Setup}
\noindent{\bf Dataset and Backbone.} We conducted the experiments on the COCO 2017 Detection Challenge tasks \cite{12}. The backbone of POST-DEYO is ResNet-50 \cite{15} pre-trained on ImageNet-1k, and our model is trained on the COCO 2017 training set without additional training data. The trained model is evaluated on the COCO 2017 validation dataset.

\noindent{\bf Implementation Details.} The POST-DEYO uses a 6-layer Transformer encoder \cite{10} and 6-layer Transformer
decoder \cite{10} with a hidden feature dimension of 256. The model is trained using the AdamW \cite{33} optimizer with a weight decay rate of 10$^-$$^4$ and a simple learning rate adjusting strategy that decreases the learning rate to 10$^-$$^5$ at epochs 11, 20, and 30. The batch size is 16, and eight GPUs with a batch size of two each are used. Additionally, the data enhancement scheme involves random cropping and scale enhancement, and the input images are resized randomly so that the short side is between 480 and 800 pixels and the long side is up to 1333 pixels. The resized images are then padded to have a 640x640 pixels resolution and are input to the first stage of the YOLO model. Two types of images with different sizes are used in the two stages of our model for the model’s forward computation. During training, the PRE-DEYO gradients are frozen.

\subsection{Main Result}
\noindent{\bf 12-epoch setting.} The progressive inference significantly
reduces the difficulty of predicting one-to-one object sets in POST-DEYO. Meanwhile, high-quality queries and anchors help POST-DEYO to clarify the learning goal, which remarkably accelerates our training process. As reported in Table~\ref{table:1}, we compare our method against strong baselines, including convolution-based methods and DETR-like methods. For a fair comparison, the FPS of all models listed in Table~\ref{table:1} are tested on the same A100 NVIDIA GPU. Note that all methods except DETR and DAB-DETR use multi-scale features. For those without multi-scale features, we report their results with ResNet-DC5 which has a better performance for its use of a dilated larger resolution feature map. Some of these methods use 5-scale feature maps, while others use 4-scale ones. We report our results using only 4 scales of feature maps.

\noindent{\bf Progressive Inference Improves Performance.} As illustrated in Table~\ref{table:2}, we compare our method against several strong baselines using the same ResNet-50 backbone. When using the common 50 epoch setting of the DETR-like schemes, our method converges faster, and thus we used 24 (2×) and 36 (3×) epoch settings. The results in Table~\ref{table:1} highlight that when using only 24 epochs, the improvement over the previous best 4-scale and 5-scale methods is 1.3 AP and 0.4 AP, respectively, and the improvement increases to 1.2 

\noindent AP and 0.9 AP for 36 epochs. The results fully demonstrate the performance improvement due to the progressive inference.
\begin{table}
\begin{center}
\begin{tabular}{cccc}
\hline
Share & PRE-Matching & Label book size & AP \\
\hline
 &\checkmark  &80 &40.2\\
\checkmark &\checkmark  &91 &40.4\\
 &\checkmark  &91 &40.9\\
\hline
\end{tabular}
\end{center}
\caption{Ablation results for DEYO using different types of Class Embedding. All models are trained under the same default settings, where share means share Embedding with CDN.}
\label{table:3}
\hspace*{\fill}
\end{table}

\begin{table}
\begin{center}
\begin{tabular}{cccc}
\hline
PRE-Matching & POST-Matching  &CDN & AP \\
\hline
 \checkmark &  & &47.7\\
 &\checkmark  &\checkmark &49.1\\
 \checkmark &\  &\checkmark &49.6\\
\hline
\end{tabular}
\end{center}
\caption{Ablation results for DEYO using different components setting. All models are trained for 12 epochs under the same default settings.}
\label{table:4}
\hspace*{\fill}
\end{table}

\begin{table}
\begin{center}
\begin{tabular}{lccccc}
\hline
PRE-DEYO & &Queries  & &AP\\
\hline
YOLOv5n & &100+800 & &47.6\\
YOLOv5x & &100+800 & &50.6\\
\hline
\end{tabular}
\end{center}
\caption{Ablation results for DEYO using different components setting. All models are trained for 12 epochs under the same default settings.}
\label{table:5}
\hspace*{\fill}
\end{table}

\subsection{Ablation Study}
In Table~\ref{table:3} and~\ref{table:4}, we utilize PRE-Matching to shield CDNs from the impact of increasing dichotomous matching stability on performance. In Table~\ref{table:3}, we explore the impact of different Class Embedding approaches on performance of the first epoch and find that using Label Embedding on CDN is most beneficial for POST-DEYO to learn information from PRE-DEYO.

We also analyze the gain of CDN on the performance of POST-DEYO, with the results in Table~\ref{table:4} revealing that CDN can guide the model to learn to correctly reconstruct the information from PRE-DEYO, in addition to increasing the stability of post-matching dichotomous matching.
In Figure~\ref{fig:4}(a), we analyze the training convergence curves using 900 queries, which are all from PRE-DEYO with different IoU thresholds for the transition components. The results demonstrate that the current filtering discrimination capability of POST-DEYO is limited, and the performance decreases as the IoU threshold rises.

We analyze the 12 epoch training convergence curves mix and raw queries, in Figure~\ref{fig:4}(b). The results show that the DEYO model using mix queries is the least using sensitive to changes in the number of queries. The DEYO model uses less computationally expensive predictions from PRE-DEYO to reduce the number of computationally expensive queries without reducing accuracy and speeding up inference.

In Table~\ref{table:5}, we analyze the impact of using different PRE-DEYO on the overall performance of DEYO. The results show that the quality of queries and anchor determines the final performance to a large extent. This is because good queries and anchor quality can establish clear optimization goals for POST-DEYO and make it easier to predict one-to-one object sets, while low-quality queries and anchor can make POST-DEYO training more difficult.

\subsection{Analysis}
Contrastive DeNoising Training is crucial for DEYO models, enhancing the POST-DEYO dichotomous matching stability and guiding POST-DEYO to obtain better results in inference. As in Let's think step by step, a well- designed intermediate inference step can substantially improve the model’s final performance. Thus, we believe that the discriminative filtering power of POST-DEYO does not only depend on the decoder but is also closely related to the design of components like CDN. In Table~\ref{table:3}, the performance degradation due to shared embedding is compared against independent embedding, suggesting that the CDN query, anchor, and the query, anchor from the POST-DEYO initialization, are ambiguous during the training. We believe that a better "CDN" and intermediate inference guidance can help POST-DEYO to exploit the potential performance of PRE-DEYO to a greater extent.

\begin{figure*}[t]
\begin{center}
\includegraphics[width=0.85\textwidth]{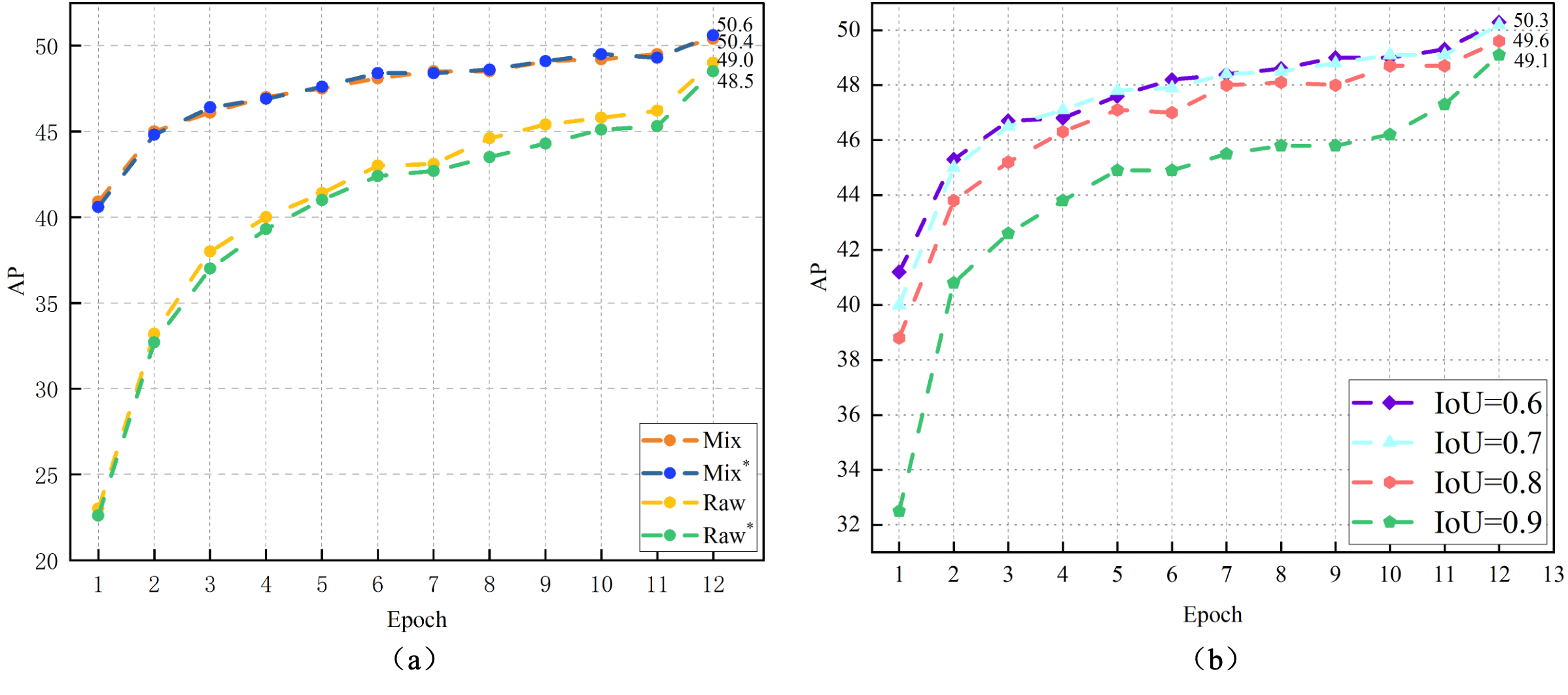}
   \caption{(a) The transition component of DEYO uses convergence curves with different IoU thresholds, where the model uses 900 queries all from PRE-DEYO. (b) DEYO's training convergence curves for using different types of queries, where Mix means the proposal of PRE-DEYO is mixed with the original POST-DEYO query, and Raw means the proposal of PRE-DEYO is not used, and all the initialized queries of the original POST-DEYO are used. Default uses 300 queries, $*$ means use 900 queries.
}
\end{center}
\label{fig:4}
\end{figure*}

The performance degradation caused by using YOLOv5n as PRE-DEYO in Table~\ref{table:5} shows that an effective Step-by-Step Object Detector is not simply a random combination to achieve superior performance. We need to carefully consider the reasonable combination of PRE-DEYO and POST-DEYO to achieve the purpose of speed and performance tradeoff.

Table~\ref{table:1} for fairness in testing speed we include the computation time of the transition component in the CPU, and in practical reasoning the computation of the DINO backbone can be executed in parallel with the computation time of the transition component in the CPU, and with a reasonable allocation of computational resources we can obtain performance gains at a very small computational cost. Also, DEYO has the advantage of using detectors like YOLO that iterate fast, while avoiding the problems with NMS in crowded scenes.

We note that Deformable-DETR also employs a similar technique to DEYO, i.e. generating proposal queries in the encoder to feed into the decoder. In fact we can achieve similar results by using two DINO cascades, but this is computationally very inefficient and does not solve any problem. This is the biggest advantage of DEYO over Deformable-DETR, i.e., it uses a more computationally efficient convolutional network and provides several orders of magnitude more queries than Deformable-DETR. We believe that the combination of one-to-many and one-to-one, and the transition from dense to sparse queries should be the paradigm for future efficient detectors, and we believe this is the key to lightweight query-based detectors in the future.

\section{Conclusion}
This paper presents a novel two-stage object detection model named DEYO, which adopts a progressive inference method based on the Step-by-Step idea. The model reduces
the difficulty of predicting one-to-one object sets for DETR-like models and solves the problem of slow convergence of the DETR-like model in a new perspective. Meanwhile, it effectively ameliorates the performance bottleneck problem of classic detectors due to NMS post- processing. The results demonstrate that the progressive inference approach significantly speeds up the convergence and improves the performance, with the best results obtained in the 1x (epochs) setting, using ResNet- 50 as the backbone.

Considering limitations and future works, the lightweight transition component and POST-DEYO have not fully exploited the information from the first stage. Indeed, Figure~\ref{fig:4}(a) highlights that the performance degradation is incurred due to inappropriately filtering the information. Hence, a more efficient information transmission scheme between the two stages should be explored, and the imperfect information encoding and decoding should be improved to avoid the final performance degradation. The two stages of our model use different backbone networks, which affect inference efficiency and the final performance. Thus, a new, dedicated denoising group may be helpful to guide the POST-DEYO model to process the information from the PRE-DEYO model. In the future, the above considerations will assist in designing a two-stage lightweight detector based on a progressive inference that can better balance the performance and speed and make information transmission more efficient.

{\small
\bibliographystyle{ieee_fullname}
\bibliography{egbib}
}

\end{document}